\documentclass[11pt]{article}

\usepackage[preprint]{acl}

\usepackage{float}    % enables [H] for tables
\usepackage{fvextra}  % enables breaklines in Verbatim
\usepackage{times}
\usepackage{latexsym}
\usepackage{amsmath}
\usepackage{svg}
\usepackage{booktabs}
\usepackage{graphicx}
\usepackage{xcolor}
\usepackage{xspace}
\usepackage[most]{tcolorbox}
\usepackage{fancyvrb}
\newcounter{promptbox}
\definecolor{bestgreen}{RGB}{5, 57, 38}
\definecolor{secondgreen}{RGB}{60,150,60}
\newcommand{\best}[1]{\textcolor{bestgreen}{\textbf{#1}}}
\newcommand{\second}[1]{\textcolor{secondgreen}{\underline{#1}}}
\newcommand{\molerag}{\textsc{MolE-RAG}\xspace}

\usepackage[T1]{fontenc}
\usepackage[utf8]{inputenc}

\usepackage{microtype}

\usepackage{inconsolata}

\usepackage{graphicx}
\usepackage{enumitem}

\title{\textbf{\molerag}: Molecular-structure Enhanced Retrieval-Augmented Generation for Chemistry}

\author{
\textbf{Joey Chan\textsuperscript{1*}, Wonbin Kweon\textsuperscript{1}, Ashley Shin\textsuperscript{2}, Niharika Bhattacharjee\textsuperscript{1},} \\
\textbf{Patrick Jiang\textsuperscript{1}, Yue Guo\textsuperscript{1}, Jiawei Han\textsuperscript{1*}} \\ \\
\textsuperscript{1}University of Illinois Urbana-Champaign \\
\textsuperscript{2}University of California, San Diego \\ \\
\texttt{\{jchan51, hanj\}@illinois.edu} \\ \\
}

\begin{document}
\maketitle
\begin{abstract}
Large language models (LLMs) have shown potential for molecular property prediction, but their ability to reason over chemical structures remains limited because molecular representations such as SMILES differ fundamentally from the natural language on which LLMs are primarily trained. To address this semantic and knowledge gap, we propose Molecule-Centric Retrieval-Augmented Generation (\molerag), a training-free framework for LLM-based molecular property prediction. \molerag augments each prediction with three complementary sources of inference-time context: retrieved chemistry literature, molecule-specific context (compound synonyms and identifiers, functional group annotations, and physicochemical descriptors), and structurally similar molecules. We evaluate \molerag across nine molecular property prediction tasks using proprietary, chemistry-specialized, and open-source LLMs. Across general-purpose LLMs, \molerag improves ROC-AUC by up to 28 points on classification tasks and reduces regression RMSE by up to $67\%$ relative to a SMILES-only baseline. We further find that context source utility varies across models and tasks, with some models benefiting more from textual retrieval, molecular context, or structural retrieval.
\end{abstract}

\section{Introduction}
Molecular property prediction has become increasingly important for accelerating drug discovery by reducing reliance on costly experimental testing \citep{shen2019molecular}. Molecular properties such as toxicity, solubility, permeability, and biological activity are central to drug discovery, as they inform key aspects of candidate quality, including safety, absorption, developability, and therapeutic potential \citep{segall2014addressing, lipinski1997experimental, schenone2013target}. Accurate prediction of these properties can help prioritize promising drug candidates, reduce downstream attrition, and improve the efficiency of molecular screening \citep{schneider2018automating}.

\begin{figure}
    \centering
    \includegraphics[width=1\linewidth]{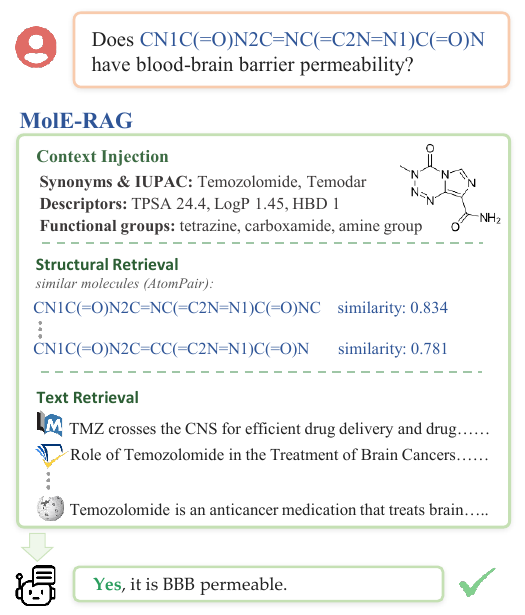}
    \caption{The \molerag framework illustrated on the BBBP task. 
    Each prediction is augmented with retrieved text passages, structurally 
    similar labeled molecules, and molecule-specific descriptors.}
    \label{fig:MolE-RAG}
\end{figure}

Retrieval-augmented generation (RAG) offers a potential way to address this limitation by providing external evidence at inference time. First, textual retrieval can supply relevant chemical and biomedical knowledge from scientific literature, helping connect molecular inputs to task-specific terminology, known biological mechanisms, and prior experimental evidence. Second, beyond retrieval, molecule-specific context, including structured physicochemical descriptors from cheminformatics toolkits, natural language molecule identifiers, and functional group annotations, can provide information that LLMs may not reliably infer from SMILES alone \citep{landrum2013rdkit, guo2023can, tang2025molfcl}. Third, structure-based retrieval is motivated by the long-standing use of molecular similarity to relate chemical structure to biological activity, ADME/Tox behavior, and physicochemical properties \citep{hendrickson1991concepts, bender2004molecular}. Recent molecular RAG approaches retrieve structurally similar molecules as contextual examples for LLM-based property prediction \citep{xian2025molrag}. Yet it remains unclear whether textual retrieval, molecular context, and structural retrieval consistently improve prediction across diverse molecular property tasks, or how different LLMs behave under these settings.

To address this gap, we propose \molerag, a Molecule-Centric Retrieval-Augmented Generation framework for LLM-based molecular property prediction. \molerag integrates three forms of inference-time context: 1) BM25-based textual retrieval \citep{robertson2009probabilistic} using LLM-augmented queries with molecule synonyms, IUPAC names, and task-specific vocabulary; 2) molecular context injection, which adds molecule identifiers, functional group annotations, and physicochemical descriptors to the prediction prompt; and 3) structural retrieval using task-adaptive molecular fingerprints to identify structurally similar training molecules. Together, these components bridge the gap between SMILES-based molecular inputs and the chemical evidence needed for property prediction. By providing relevant chemical context at inference time, \molerag also avoids additional model fine-tuning and post-training.

In this work, we make three main contributions. First, we propose \molerag, a training-free framework that augments LLM-based molecular property prediction with retrieved chemistry and biomedical literature, molecule-specific context, and structurally similar molecules. Second, we develop a task-adaptive context augmentation strategy that connects SMILES inputs to chemistry literature through LLM-mediated query augmentation, injects interpretable molecule-level identifiers, functional groups, and physicochemical descriptors, and retrieves chemically relevant examples through task-specific structural fingerprints. Third, we systematically evaluate general-purpose, chemistry-specialized, and proprietary LLMs across multiple context configurations to identify when textual retrieval, molecular context, and structural retrieval improve prediction, and how these effects vary across model families and molecular property tasks.

\section{Related Works}
% https://arxiv.org/pdf/2507.07456 page 55

% \subsection{General-Purpose Chemistry Large Language Models (GPMs)} 
% https://arxiv.org/pdf/2401.14818 

% \subsection{RAG for Medicine} 
% https://journals.plos.org/digitalhealth/article?id=10.1371/journal.pdig.0000877 
% https://www.nature.com/articles/s41746-025-01519-z

\subsection{LLMs for Chemistry}
Since the introduction of transformer-based models, there have been a several works focusing on using language models for science \cite{SPECTER, MedCPT, Zhang2024survey}. 
There has been increasing interest in using LLMs for chemistry, including recent works such as MoleculeNet \cite{moleculenet} and MolInstructions \cite{MolInstructions}. 
ChemDFM \cite{ChemDFM2025} is a recent chemistry foundational model that finetunes LLaMa-13B \cite{llama}. 
ChemDFM is claimed to outperform general LLMs such as GPT-4 \cite{gpt4} and LLaMa-2 \cite{llama}, but it was only evaluated on ChemLLMBench, which is severely limited as its test set is limited to 100 samples per task \cite{ChemLLMBench}.

\subsection{RAG for Property Prediction} 
Molecular property prediction plays an important role in computational chemistry. By identifying candidate molecules with desired pharmacological properties, accurate molecular property prediction can accelerate drug discovery \cite{xia2023understanding, walters2020applications}. Retrieval-Augmented Generation (RAG) has been especially promising because it can leverage the latest advances in general-purpose LLMs without requiring further training. Recent works such as ChemRAG \cite{Zhong-ChemRAG} and MolRAG \cite{xian-molrag} show the promise of RAG for property prediction, as in-context learning from retrieved examples greatly improve performance. 

\begin{figure*}
    \centering
    \includegraphics[width=1\linewidth]{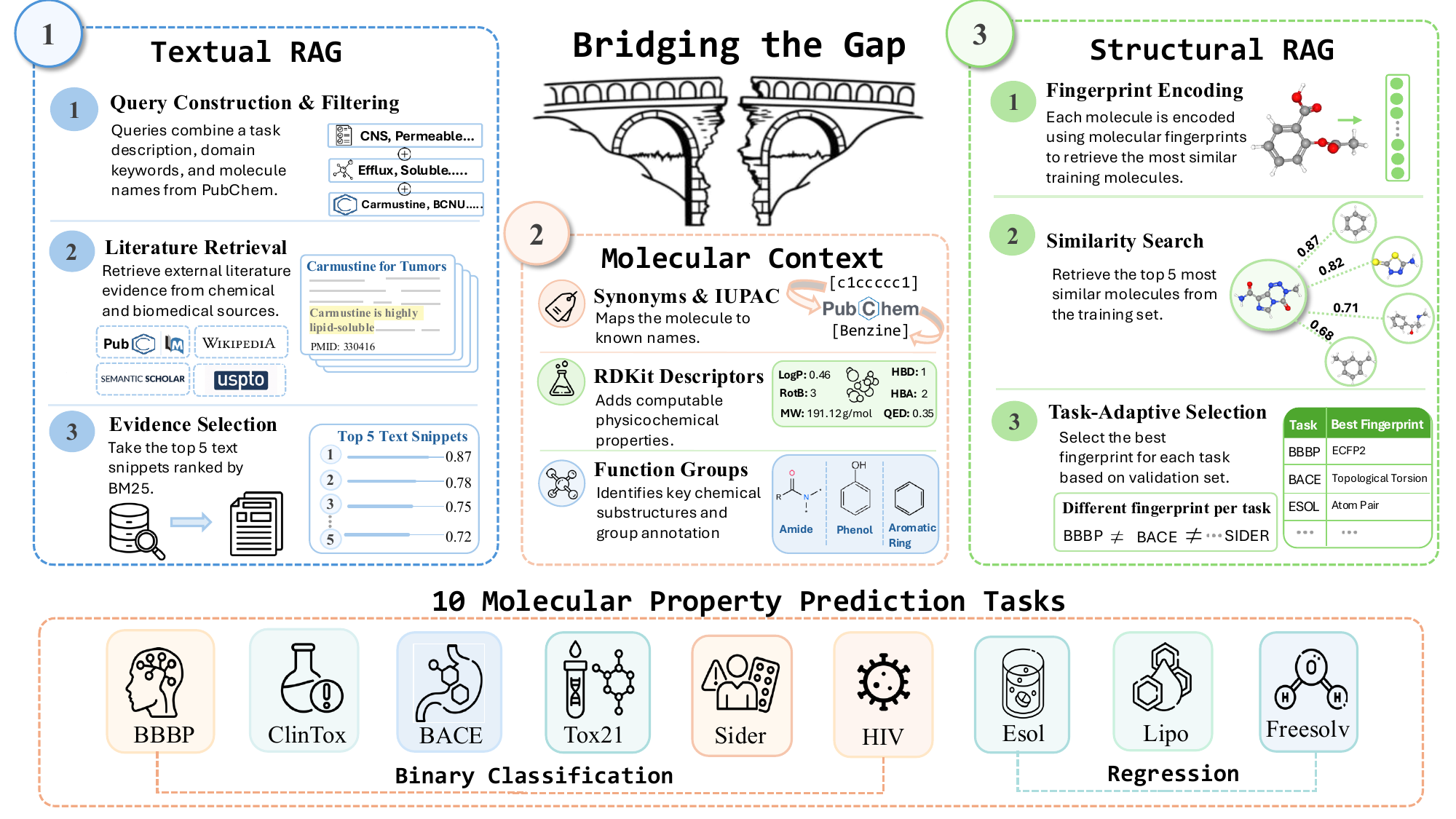}
    \caption{\textbf{The \molerag Framework.} Three complementary sources of inference-time context augment each prediction: (1) \textbf{Text retrieval} 
    constructs a hybrid query from the task description, LLM-generated domain keywords, and filtered molecule names, retrieving the top-5 passages from the ChemRAG corpus \citep{zhong2025benchmarking}; (2) \textbf{Molecular Context} appends compound identifiers, task-adaptive RDKit descriptors, and functional group annotations; and (3) \textbf{Structure retrieval} encodes each molecule using a task-specific molecular fingerprint 
    and retrieves the top-5 most similar training molecules as labeled few-shot examples, where the best fingerprint per task is selected on the validation set. \molerag is evaluated on nine molecular property prediction tasks.}
    \label{fig:MolE-RAG_pipeline}
\end{figure*}

\section{Method}
\molerag is a training-free RAG framework for LLM-based molecular property prediction. Instead of relying on a single source of external context, \molerag augments each prediction with three complementary forms of inference-time evidence: retrieved chemistry literature, structurally similar molecules, and molecule-specific descriptors. Figure~\ref{fig:MolE-RAG_pipeline} illustrates the overall framework.

\subsection{Problem Formulation}
Let $D = \{(M_i, y_i)\}_{i=1}^{N}$ be a training set of labeled molecules, where each molecule $M_i$ is represented as a SMILES string with an associated property label $y_i$. Given a query molecule $M$, the goal of molecular property prediction is to estimate its task-specific property $\hat{y} = f(M)$, where $\hat{y}$ is either a binary label for classification tasks or a continuous value for regression tasks.

In \molerag, multiple sources of inference-time context can be used to augment the input molecule. The textual retriever $R_{\text{text}}$, implemented using BM25, retrieves relevant passages from an external chemistry corpus $K$. The molecular context component derives molecule-specific context directly from $M$, including molecule synonyms and identifiers, functional group annotations, and RDKit physicochemical descriptors. Finally, the structural retriever $R_{\text{struct}}$ performs structure-based retrieval by identifying structurally similar labeled molecules from the training set $D$. These three context sources are defined as:
\[
\begin{aligned}
C_{\text{text}} &= R_{\text{text}}(M \mid K), \\
C_{\text{mol}} &= g_{\text{mol}}(M), \\
C_{\text{struct}} &= R_{\text{struct}}(M \mid D).
\end{aligned}
\]

Since different LLMs and molecular property tasks may benefit from different types of context, we allow these sources to be used individually or in combination. Let $S \subseteq \{\text{text}, \text{mol}, \text{struct}\}$ denote the selected context configuration for a given experiment. The corresponding augmented context is defined as:
\[
C_S = \{C_j : j \in S\},
\]
where $C_{\text{text}}$, $C_{\text{mol}}$, and $C_{\text{struct}}$ correspond to textual evidence, molecule-specific context, and structurally similar molecules, respectively. The augmented prompt is then constructed as:
\[
P_S = (I, M, C_S),
\]
where $I$ denotes the task instruction. Finally, the LLM $L$ produces the prediction:
\[
\hat{y} = L(P_S).
\]
This formulation allows us to evaluate not only whether these inference-time context sources improve molecular property prediction, but also how the usefulness of each context source varies across models, tasks, and prediction settings.

\subsection{Textual Retrieval}
Given a query molecule $M$, the textual retriever $R_{\text{text}}$ retrieves relevant passages from an external chemistry corpus $K$. We implement $R_{\text{text}}$ using BM25 \citep{robertson2009probabilistic} over the ChemRAG corpus \citep{zhong2025benchmarking}, which aggregates text snippets from PubChem, PubMed, USPTO, Semantic Scholar, Wikipedia, and OpenStax textbooks. These sources span molecular databases, biomedical literature, chemical patents, and educational resources. Because BM25 relies on lexical term matching, retrieval quality depends strongly on constructing queries with chemically meaningful terms that overlap with relevant passages.

A key challenge is that SMILES strings are poor lexical queries: they encode molecular structure but share little vocabulary with scientific literature. Therefore, \molerag uses a task-aware hybrid query that combines the prediction objective, task-specific terminology, and molecule identity:
\[
q_{\text{hybrid}} =
\bigl(q_{\text{task}}, q_{\text{keywords}}, q_{\text{mol}}\bigr).
\]
Here, $q_{\text{task}}$ denotes the natural language task description, $q_{\text{keywords}}$ denotes task-specific chemical terminology, and $q_{\text{mol}}$ denotes molecule identity information, such as PubChem synonyms, the IUPAC name, or the SMILES string.

The task description states the prediction objective in natural language, such as \textit{``whether the molecule has blood-brain barrier permeability.''} Task keywords are generated once per task using GPT-4o-mini \citep{openai2024gpt4omini} and cached for reuse, producing domain-relevant terms such as assay names, pathway terminology, mechanistic keywords, and property-specific synonyms. The molecule identity information is selected from PubChem synonyms when available, with fallback to the IUPAC name or SMILES string.

Together, these components increase lexical overlap with relevant chemistry passages while avoiding noisy identifiers or uninformative tokens. The top-$k$ passages retrieved using $q_{\text{hybrid}}$, with $k=5$ in our experiments, are prepended to the prediction prompt as textual evidence.

\begin{table*}[t]
\centering
\small
\setlength{\tabcolsep}{3.0pt}
\makebox[\textwidth][c]{
\begin{tabular}{l|cccccc|ccc}
\toprule
& \multicolumn{6}{c|}{\textbf{Classification}} &
\multicolumn{3}{c}{\textbf{Regression}} \\
\cmidrule(lr){2-7}
\cmidrule(lr){8-10}
\textbf{Dataset}
& BBBP & Tox21 & SIDER & ClinTox & BACE & HIV
& ESOL & FreeSolv & Lipophilicity \\
\midrule
\# Molecules
& 2039 & 7831 & 1427 & 1478 & 1513 & 41127
& 1128 & 642 & 4200 \\
\# Tasks
& 1 & 12 & 27 & 2 & 1 & 1
& 1 & 1 & 1 \\
\midrule
MGCN
& 85.0 & 70.7 & 55.2 & 63.4 & 73.4 & ---
& 1.266 & 3.349 & 1.113 \\
GROVER
& 84.1 & 80.0 & 58.2 & 70.7 & 83.3 & 67.5
& 1.475 & 3.235 & 0.974 \\
MolCLR$_{\mathrm{GCN}}$
& 73.2 & 73.5 & 61.0 & 85.9 & 82.6 & 76.8
& 1.102 & 2.241 & 0.819 \\
MolCLR$_{\mathrm{GIN}}$
& 73.5 & 76.7 & 60.7 & 90.4 & 83.5 & 77.6
& 1.091 & 2.017 & 0.824 \\
MolCLR$_{\mathrm{CMPNN}}$
& 72.4 & 78.4 & 59.7 & 88.0 & 85.0 & 77.8
& 0.911 & 2.021 & 0.875 \\
SchNet
& 84.8 & 76.6 & 54.5 & 71.7 & 76.6 & ---
& 1.045 & 3.215 & 0.909 \\
\midrule
\textbf{Llama$_{\text{3.2-3B}}$} & & & & & & & & & \\
\quad Baseline
& 48.2 & 50.0 & 49.9 & 48.3 & 50.4 & 51.8
& 3.175 & 5.995 & 4.095 \\
\quad $+\molerag$
& 48.3 & 52.5 & 50.3 & 48.8 & 51.8 & 58.7
& 2.181 & 4.343 & 1.706 \\
\textbf{Mistral$_{\text{7B-Inst}}$} & & & & & & & & & \\
\quad Baseline
& 46.1 & 53.1 & 50.4 & 50.0 & 48.6 & 45.9
& 3.474 & 12.585 & 3.347 \\
\quad $+\molerag$
& 74.6 & 54.8 & 50.8 & 62.1 & 76.8 & 62.4
& 2.890 & 4.128 & 1.210 \\
\textbf{Qwen3$_{\text{4B-Inst}}$} & & & & & & & & & \\
\quad Baseline
& 53.0 & 49.6 & 51.1 & 50.3 & 48.9 & 53.3
& 2.537 & 5.524 & 2.456 \\
\quad $+\molerag$
& \best{80.1} & 66.2 & 52.8 & 56.6 & 71.1 & \best{73.5}
& 1.852 & \second{2.700} & 1.105 \\
\textbf{ChemDFM$_{\text{14B}}$} & & & & & & & & & \\
\quad Baseline
& 79.0 & 55.0 & 49.2 & 49.7 & 68.2 & 55.2
& 4.086 & 6.523 & 2.347 \\
\quad $+\molerag$
& \second{78.1} & \best{68.8} & \second{53.9} & 59.4 & \best{80.4} & \second{69.2}
& 1.467 & 3.476 & 1.021 \\
\textbf{GPT-4o-mini} & & & & & & & & & \\
\quad Baseline
& 54.9 & 52.3 & 51.6 & 52.0 & 51.3 & 60.0
& 2.835 & 5.475 & 1.667 \\
\quad $+\molerag$
& 74.7 & 59.3 & 53.5 & \best{63.3} & 57.4 & 61.3
& \second{1.171} & \best{2.555} & \second{0.964} \\
\textbf{GPT-5.4-nano} & & & & & & & & & \\
\quad Baseline
& 53.4 & 52.7 & 51.6 & 52.8 & 54.3 & 50.5
& 1.723 & 5.851 & 1.448 \\
\quad $+\molerag$
& 77.0 & \second{67.0} & \best{56.6} & \second{62.7} & \second{78.3} & 65.1
& \best{1.138} & 2.880 & \best{0.894} \\
\bottomrule
\end{tabular}
}
\caption{Molecular Property Prediction Benchmark Results. Performance on six classification benchmarks (ROC-AUC) and three regression benchmarks (RMSE). Baselines reported in \cite{fang2023knowledge} were included in our evaluation. \best{Bold dark green} marks the best \molerag-augmented model per task; \second{underlined light green} marks the second-best.}
\label{tab:model_performance}
\end{table*}

\subsection{Molecular Context}
Beyond textual and structure-based retrieval, \molerag directly injects molecule-specific context into the prediction prompt, producing $C_{\text{mol}}$ without requiring additional external search. This addresses a limitation of SMILES-only inputs: although SMILES encode molecular topology, they do not explicitly expose chemically meaningful information such as compound identity, functional groups, or physicochemical properties. $C_{\text{mol}}$ consists of three components.

\paragraph{Compound identifiers}
We first inject the molecule's common drug name, up to five additional synonyms, and IUPAC name, sourced from PubChem. These identifiers are selected from the same synonym cache used by $R_{\text{text}}$, with catalog identifiers and registry numbers removed. This grounds the prompt in the molecule's real-world identity, which may help the LLM connect the compound to chemical or pharmacological knowledge encountered during pretraining.

\paragraph{Functional group annotations}
We detect functional groups using AccFG \citep{liu2025accfg} and inject them as a concise natural language statement, such as \textit{``Functional groups present: carboxylic acid, ester, aromatic ring.''} Functional groups provide chemically meaningful substructures that are relevant to reactivity, toxicity, and ADME-related properties, but may not be obvious to an LLM from the SMILES string alone.

\paragraph{Physicochemical descriptors}
We compute RDKit descriptors \citep{landrum2013rdkit} directly from the SMILES string and inject task-relevant descriptors as a structured block. Rather than including all available RDKit descriptors, which may introduce redundant, weakly relevant, or overly verbose information into the prompt, we select a compact task-specific subset. For each task, descriptors are ranked by their absolute correlation with the training labels, using point-biserial correlation for classification tasks and Pearson correlation for regression tasks, and the top 15 descriptors are retained. This keeps the prompt focused on descriptors most associated with the target property while reducing unnecessary context length.

When applicable, descriptors are accompanied by qualitative labels, such as \textit{``MolLogP: 2.1 (moderate lipophilicity).''} We also append Lipinski's Rule of 5 compliance, which summarizes widely used physicochemical criteria related to oral absorption and permeability, including molecular weight, lipophilicity, hydrogen-bond donors, and hydrogen-bond acceptors \citep{lipinski1997experimental}. This provides a broadly interpretable drug-likeness signal that is relevant across many small-molecule property prediction tasks.

To assess the contribution of each molecular context component, we ablate molecule synonyms and identifiers, functional group annotations, and physicochemical descriptors in Appendix~\ref{app:rdkit}.

\subsection{Structure-Based Retrieval}
Given a query molecule $M$, the structure-based retriever $R_{\text{struct}}$ retrieves the top-$k$ most structurally similar molecules from the training set $D$. We first encode the query molecule and each training molecule $M_i \in D$ using a molecular fingerprint function $\phi(\cdot)$. The similarity between $M$ and $M_i$ is then computed using Tanimoto similarity:
\[
s_i = \mathrm{Tanimoto}\bigl(\phi(M), \phi(M_i)\bigr).
\]
The structure-based context is defined as the top-$k$ labeled molecules with the highest similarity scores:
\[
C_{\text{struct}} =
\operatorname{TopK}_{(M_i,y_i)\in D}(s_i, k).
\]
These retrieved molecules are used as labeled few-shot examples in the prediction prompt. We set $k=5$ for all experiments. This retrieval strategy is motivated by the principle that structurally similar molecules often exhibit similar biological or physicochemical properties \citep{bender2004molecular}.

We evaluate structure-based retrieval across nine molecular property prediction tasks from MoleculeNet \citep{wu2018moleculenet}, covering six binary classification tasks and three regression tasks. For each task, the training set serves as the retrieval database $D$, while the validation set is used to select the fingerprint type that yields the best retrieval performance.

Molecular fingerprints encode chemical structures as fixed-length vectors for similarity search \citep{capecchi2020one}. We consider several fingerprint families that capture complementary aspects of molecular structure. Circular fingerprints, including ECFP2, ECFP4, and ECFP6, encode local atom neighborhoods at increasing radii using the Morgan algorithm \citep{rogers2010extended}. Functional-class fingerprints, including FCFP4 and FCFP6, represent pharmacophoric atom roles rather than exact atom identities, allowing similarity to reflect shared functional behavior. MACCS keys encode molecules using a fixed set of predefined structural patterns, while topological fingerprints such as Atom Pair and Topological Torsion capture connectivity patterns and longer-range relationships within the molecular graph \citep{cereto2015molecular}. 

Since different fingerprint representations encode distinct structural information and may perform differently across molecular property prediction tasks \citep{xie2020improvement}, \molerag does not assume a single fingerprint is optimal across all tasks. Instead, it selects the best-performing fingerprint on the validation set for each task. The selected fingerprint for each task is reported in Appendix~\ref{app:fingerprints}.

\subsection{Prompt Design}
For each prediction setting, the selected context sources are assembled into a structured prompt. When included, context sources are presented in a fixed order: task instruction, retrieved text passages, molecule-specific context, structurally similar labeled molecules, and the query molecule. This ordering provides broad chemical background first, then molecule-level information, followed by structurally similar examples before the final query.

The output instruction is adapted to the task type, requiring either a categorical prediction for classification tasks or a continuous value for regression tasks.

\section{Experiment}
\subsection{Experiment Settings}
\paragraph{Dataset splits}
We use scaffold splitting with three random seeds and an 8:1:1 training/validation/test ratio across all datasets, following previous studies \citep{rong2020self, fang2023knowledge}. The training set is used as the retrieval pool for structure-based retrieval, and the validation set is used for task-adaptive fingerprint selection and molecular descriptor selection. 

\paragraph{Models}
We evaluate \molerag across LLMs from three categories: proprietary models (GPT-4o-mini\citep{openai2024gpt4omini} and GPT-5.4-nano \citep{singh2025openai}); general-purpose open-source models (Llama-3.2-3B-Instruct, Mistral-7B-Instruct-v0.3, and Qwen3-4B-Instruct \citep{grattafiori2024llama, jiang2023mistral, yang2025qwen3}); 
and the chemistry-specialized model ChemDFM-v2.0-14B \citep{ChemDFM2025}. All models are evaluated at temperature $0$ in a zero-shot setting to ensure reproducibility and isolate the contribution of retrieval augmentation. This diverse model set allows us to examine how model capability and chemistry-specific finetuning affect the benefit of retrieval and molecular context augmentation.

\subsection{Baselines}
We compare \molerag against a representative set of supervised and self-supervised graph baselines on the MoleculeNet benchmarks. Supervised graph neural networks include SchNet~\citep{schutt2017schnet} and MGCN~\citep{lu2019molecular}, which learn directly from molecular graphs under task supervision. Self-supervised pretraining methods include GROVER~\citep{rong2020self} and MolCLR~\citep{wang2022molecular}, the latter evaluated across three GNN backbones (GCN, GIN, CMPNN). Reported baseline numbers are taken from \citet{fang2023knowledge}. Our aim is to characterize the contribution of retrieval and molecular context augmentation to LLM-based prediction rather than to claim state-of-the-art on MoleculeNet, so stronger knowledge-enhanced graph methods such as KANO~\citep{fang2023knowledge} and GODE~\citep{jiang2025bi} are discussed for context but not included as headline comparators.

\subsection{Metrics}
For classification tasks, we report the mean ROC-AUC across three random seeds. For regression tasks, we report the mean RMSE across three random seeds.

\subsection{Effect of \molerag on Classification Tasks}
We evaluate \molerag on six binary classification tasks: BBBP, BACE, ClinTox, HIV, Tox21, and SIDER. The SMILES-only baseline provides only the molecule SMILES string and task description, while \molerag augments the prompt with retrieved text, structurally similar labeled molecules, and molecule-specific descriptors. Results are reported as ROC-AUC in Table~\ref{tab:model_performance}.

Across general-purpose LLMs, \molerag improves ROC-AUC over the SMILES-only baseline on nearly every dataset, with the largest gains appearing for models whose SMILES-only performance is close to random. Mistral improves on BBBP from 46.1 to 74.6 and on BACE from 48.6 to 76.8 --- gains of more than 28 ROC-AUC points. Qwen3 improves on BBBP from 53.0 to 80.1 and on HIV from 53.3 to 73.5. These results suggest that inference-time molecular context can substantially compensate for the limited chemical information available from SMILES-only prompting.

Without additional context, smaller open-source models generally trail proprietary models on classification. With \molerag, however, Mistral-7B and Qwen3-4B become competitive with --- and on several tasks outperform --- the proprietary baselines. Qwen3 + \molerag obtains the best result among all evaluated LLMs on BBBP (80.1) and HIV (73.5). These findings indicate that access to relevant molecular context is a major bottleneck for smaller LLM-based property prediction, and that retrieval can largely close the gap with stronger proprietary models.

The gains are not limited to smaller open-source models. GPT-5.4-nano + \molerag is consistently among the top two methods on Tox21, SIDER, ClinTox, and BACE, taking the top score on SIDER (56.6). ChemDFM-14B + \molerag, the only chemistry-specialized model in our evaluation, achieves the highest ROC-AUC on Tox21 (68.8) and BACE (80.4), and is second-best on BBBP, SIDER, and HIV. The exception is Llama-3.2-3B, which shows only marginal gains (BBBP from 48.2 to 48.3, ClinTox from 48.3 to 48.8) and remains near random on most classification tasks, suggesting that effective use of retrieved chemical context requires sufficient model capacity.

Compared with the supervised and self-supervised graph baselines, \molerag-augmented LLMs reach competitive but not state-of-the-art performance on classification. On BBBP, Qwen3 + \molerag (80.1) trails the strongest graph baselines (MGCN at 85.0, SchNet at 84.8) by roughly five ROC-AUC points but exceeds the entire MolCLR family (72.4--73.5). On BACE, ChemDFM + \molerag (80.4) approaches GROVER (83.3) and the MolCLR family (82.6--85.0). On HIV, Qwen3 + \molerag (73.5) is within four points of the strongest graph baseline (MolCLR$_{\mathrm{CMPNN}}$ at 77.8) and clearly exceeds GROVER (67.5). The largest remaining gap is on ClinTox, where the best \molerag result (GPT-4o-mini at 63.3) lies more than twenty points below MolCLR$_{\mathrm{GIN}}$ (90.4) --- a task that appears to benefit most from explicit graph-level structure.

\subsection{Effect of \molerag on Regression Tasks}
We further evaluate \molerag on three regression tasks: ESOL, FreeSolv, and Lipophilicity. Across all evaluated LLMs, \molerag consistently lowers regression error relative to the SMILES-only baseline. The largest reductions occur on FreeSolv, where several baselines produce highly inaccurate predictions: Mistral improves from RMSE 12.585 to 4.128 --- a 67\% reduction --- and Qwen3 from 5.524 to 2.700. \molerag also yields substantial improvements on ESOL (ChemDFM from 4.086 to 1.467, GPT-4o-mini from 2.835 to 1.171) and on Lipophilicity (Llama from 4.095 to 1.706, Mistral from 3.347 to 1.210).

Unlike classification, no single model dominates all regression tasks. GPT-5.4-nano + \molerag achieves the best result on ESOL (1.138) and Lipophilicity (0.894), while GPT-4o-mini + \molerag is best on FreeSolv (2.555) and second-best on the other two. This variation suggests that different molecular properties benefit from different combinations of model capability and retrieved context, rather than a single uniformly optimal LLM.

Compared with the graph baselines, \molerag-augmented LLMs are more competitive on regression than on classification. The best \molerag results outperform MGCN and GROVER on all three regression tasks, and outperform SchNet on FreeSolv and Lipophilicity. On ESOL, GPT-5.4-nano + \molerag (1.138) outperforms MGCN (1.266) and GROVER (1.475) but trails SchNet (1.045) and the MolCLR family (0.911--1.102). On FreeSolv, GPT-4o-mini + \molerag (2.555) outperforms MGCN (3.349), GROVER (3.235), and SchNet (3.215), trailing only the MolCLR family (2.017--2.241). On Lipophilicity, GPT-5.4-nano + \molerag (0.894) is the lowest RMSE outside the MolCLR family (0.819--0.875), beating SchNet (0.909), GROVER (0.974), and MGCN (1.113). The MolCLR family remains the strongest graph baseline overall but is approached most closely on the regression benchmarks, where targeted descriptor injection appears to be more decisive than message passing.

A second benefit of \molerag is that it reduces extreme numeric failures from SMILES-only prompting. Several baselines produce regression outputs far outside the expected range, especially on FreeSolv, where Mistral's SMILES-only RMSE of 12.585 is nearly twice that of the next-worst model. Adding retrieved examples and molecule-specific descriptors appears to calibrate the prediction space, reducing both average error and the incidence of extreme outliers. Overall, the regression results show that \molerag not only improves average accuracy but also makes LLM-based numeric prediction more stable.

\subsection{Ablation Variants}
To measure the contribution of each retrieval source, we compare \molerag against a zero-shot baseline and two families of ablation variants. The zero-shot baseline uses only the task instruction and query SMILES. Each ablation variant isolates one form of inference-time context.

\paragraph{Textual retrieval}
Naive RAG~\citep{lewis2020retrieval, gao2023retrieval} retrieves the top-$5$ passages from the ChemRAG corpus using BM25 over the query SMILES. \molerag$_{\text{hybrid}}$ expands the BM25 query with LLM-generated task keywords and filtered molecule synonyms to test whether task-aware query construction improves retrieval. Table~\ref{tab:ablation_text} reports the comparison.

\begin{table}[!t]
\centering
\footnotesize
\setlength{\tabcolsep}{3.0pt}
\resizebox{\columnwidth}{!}{%
\begin{tabular}{l|cccccc}
\toprule
\textbf{Method}
& BBBP & BACE & ClinTox & HIV & Tox21 & SIDER \\
\midrule
Zero-shot
& 55.8 & 53.6 & 50.5 & 52.8 & 52.1 & 50.6 \\
Naive RAG
& 49.6 & 50.4 & 49.5 & 52.8 & 53.4 & 51.0 \\
\molerag$_{\text{hybrid}}$
& \best{57.9} & \best{54.0} & \best{50.2} & \best{54.6} & \best{54.6} & \best{52.2} \\
\bottomrule
\end{tabular}}
\caption{Text-retrieval ablation. ROC-AUC averaged across the six evaluated LLMs. Zero-shot and Naive RAG share the HIV value because the SMILES baseline reuses the BM25 pipeline for that task.}
\label{tab:ablation_text}
\end{table}

Naive BM25 retrieval over the raw SMILES string reduces mean ROC-AUC on BBBP, BACE, and ClinTox and matches the baseline elsewhere, indicating that the SMILES string is a poor lexical query against general chemistry text. The hybrid variant, which augments the BM25 query with LLM-generated task keywords and filtered synonyms, recovers from this regression and yields small but consistent gains over zero-shot on all six tasks. This confirms that query construction matters as much as the corpus itself when retrieving textual context for molecular tasks.

\paragraph{Structural retrieval}
\molerag$_{\text{struct}}$ retrieves the top-$5$ nearest training molecules using the best-performing molecular fingerprint for each task, selected on the validation set, and presents them as labeled in-context demonstrations. We compare this against Random few-shot, which samples $5$ training molecules uniformly at random in the same format. Table~\ref{tab:ablation_struct} reports the comparison.

\begin{table}[!t]
\centering
\footnotesize
\setlength{\tabcolsep}{3.0pt}
\resizebox{\columnwidth}{!}{%
\begin{tabular}{l|cccccc}
\toprule
\textbf{Method}
& BBBP & BACE & ClinTox & HIV & Tox21 & SIDER \\
\midrule
Zero-shot
& 55.8 & 53.6 & 50.5 & 52.8 & 52.1 & 50.6 \\
Random few-shot
& 59.7 & 55.8 & 54.3 & --- & 51.3 & 50.3 \\
\molerag$_{\text{struct}}$
& \best{73.3} & \best{73.3} & \best{59.6} & \best{69.4} & \best{65.9} & \best{53.6} \\
\bottomrule
\end{tabular}}
\caption{Structural-retrieval ablation. ROC-AUC averaged across the six evaluated LLMs. Random few-shot samples 5 training molecules uniformly; \molerag$_{\text{struct}}$ retrieves the top-5 fingerprint neighbors.}
\label{tab:ablation_struct}
\end{table}

Structural retrieval is the strongest single context source we evaluate. \molerag$_{\text{struct}}$ improves over the zero-shot baseline on every classification task --- by roughly $20$ ROC-AUC points on BBBP and BACE, and by $17$ points on HIV. Random few-shot serves as a sanity check: it yields modest gains over zero-shot on BBBP, BACE, and ClinTox, but the much larger gap between Random and \molerag$_{\text{struct}}$ shows that fingerprint-based neighbor selection contributes well beyond the in-context-learning effect of simply seeing labeled molecules.

\section{Conclusion}
In this work, we introduced \molerag and benchmarked it against representative supervised and self-supervised graph baselines (MGCN, SchNet, GROVER, and the MolCLR family) on MoleculeNet. We found that retrieval-augmented LLMs close most of the gap to these specialized graph models on regression and on three of six classification tasks, while graph-based methods remain stronger on toxicity benchmarks where graph-level structure carries the most signal. Across model scales, mid-sized open-source LLMs such as Mistral-7B and Qwen3-4B gain the most from \molerag and become competitive with proprietary baselines, while the smallest model in our evaluation, Llama-3.2-3B, shows only marginal gains, suggesting that effective use of inference-time chemical context requires sufficient model capacity. Together, these findings suggest that, equipped with the right context, training-free LLMs can serve as a practical alternative for molecular property prediction.

\section{Limitations}
A limitation of this work is that although we used the ChemRAG corpus and BM25 retriever in our ablation study, the significant computational overhead of the large ChemRAG corpus prevented us from evaluating the full ChemRAG method with dense retrievers. Additionally, we did not evaluate chain-of-thought prompting, which prior work~\citep{xian2025molrag} has shown can further improve LLM reasoning on molecular property prediction; integrating CoT with our retrieval pillars is left to future work.

\section{Acknowledgments}
This work was supported by the Molecule Maker Lab Institute: An AI Research Institutes program supported by NSF (No. 2019897, United States). Computation for this work also used Delta GPUs at NCSA through allocation [CIS240504] from the Advanced Cyberinfrastructure Coordination Ecosystem: Services \& Support (ACCESS) program, which is supported by U.S. National Science Foundation grants \#2138259, \#2138286, \#2138307, \#2137603, and \#213829.

% Bibliography entries for the entire Anthology, followed by custom entries
%\bibliography{anthology,custom}
% Custom bibliography entries only

\clearpage

\appendix

\paragraph{Appendix organization}
The appendix is organized as follows. Appendix~\ref{app:reproduce} provides reproducibility details. Appendix~\ref{app:fingerprints} reports the task-specific molecular fingerprints selected for structure-based retrieval. Appendix~\ref{app:rdkit} presents the RDKit descriptors selected for molecule-specific context injection. Appendix~\ref{app:per-task} provides additional results for multitask datasets. Appendix~\ref{app:ai-usage} describes the use of AI assistants, and Appendix~\ref{app:artifacts} summarizes the scientific artifacts used in this work. Finally, Appendix~\ref{app:prompts} presents the prompt templates used across experimental conditions.

\section{Reproducibility}
\label{app:reproduce}
We provide the source code\footnote{https://github.com/jchan58/MolE-RAG.git} and configuration for the key experiments. 

\section{Best Fingerprint Per Task}
\label{app:fingerprints}

To select the most effective structure-based retrieval representation, we performed a validation sweep over multiple molecular fingerprint types for each task. The training set was used as the retrieval pool, and each validation molecule was matched to its top-$k$ most similar training molecules under each fingerprint representation. For classification tasks, we selected the fingerprint with the highest validation ROC-AUC, using accuracy as a secondary criterion when applicable. For regression tasks, we selected the fingerprint with the lowest validation RMSE, using MAE as a secondary criterion when applicable. The selected fingerprint was then fixed and used for test-set evaluation.

The selected fingerprints varied across tasks, suggesting that no single molecular representation was uniformly optimal across all property prediction settings. Circular fingerprints such as ECFP2 were selected for BBBP, indicating that local atom-neighborhood patterns were useful for blood-brain barrier permeability prediction. In contrast, topological representations such as Topological Torsion and Atom Pair were selected for BACE, ESOL, Lipo, and Tox21, suggesting that longer-range connectivity patterns can be informative for activity, solubility, lipophilicity, and toxicity-related tasks. Functional-class fingerprints such as FCFP2 were selected for ClinTox and SIDER, indicating that pharmacophoric or functional-role information may be useful for clinical toxicity and side-effect prediction. Overall, these results support the use of task-adaptive fingerprint selection rather than relying on a single fixed structural representation for all datasets.

Table~\ref{tab:best_fingerprints} reports the selected fingerprint for each task.

\begin{table}[H]
\centering
\small
\begin{tabular}{ll}
\toprule
Dataset & Selected Fingerprint \\
\midrule
BACE & Topological Torsion \\
BBBP & ECFP2 \\
ClinTox & FCFP2 \\
ESOL & Atom Pair \\
FreeSolv & MACCS \\
HIV & RDKit \\
Lipo & Atom Pair \\
SIDER & FCFP2 \\
Tox21 & Atom Pair \\
\bottomrule
\end{tabular}
\caption{Selected structure-based retrieval fingerprints for each molecular property prediction task. Fingerprints were selected using validation-set performance and then fixed for test-set evaluation.}
\label{tab:best_fingerprints}
\end{table}

\section{Best RDKit Descriptors Per Task}
\label{app:rdkit}

For each task, we retained the top 15 RDKit descriptors ranked by absolute correlation with the target label. For classification tasks, descriptors were ranked using point-biserial correlation, while regression tasks used Pearson correlation. For multi-assay datasets, descriptor ranking was performed separately for each assay.

Across tasks, we observed several recurring descriptor patterns. Molecular size and complexity descriptors, such as \texttt{MolWt}, \texttt{HeavyAtomCount}, \texttt{BertzCT}, and Chi descriptors, appeared frequently across both classification and regression tasks, suggesting that global molecular structure provides useful signal for many property prediction settings. Polarity and hydrogen-bonding descriptors, including \texttt{TPSA}, \texttt{NOCount}, \texttt{NumHDonors}, \texttt{NumHAcceptors}, and \texttt{NHOHCount}, were especially prominent for BBBP and FreeSolv, which is consistent with the importance of polar surface area and hydrogen-bonding capacity for permeability and solvation-related properties. Lipophilicity and surface-area descriptors, such as \texttt{MolLogP}, \texttt{SlogP\_VSA}, and \texttt{SMR\_VSA}, also appeared across multiple tasks, reflecting the role of hydrophobicity and surface-area partitioning in molecular activity, toxicity, solubility, and lipophilicity. Finally, several functional-group descriptors, including \texttt{fr\_azo}, \texttt{fr\_quatN}, \texttt{fr\_aniline}, \texttt{fr\_phenol}, and \texttt{fr\_COO}, were highly ranked for specific datasets, indicating that task-specific substructures provide complementary signal beyond general physicochemical descriptors.

Tables~\ref{tab:rdkit_classification_descriptors_a} and~\ref{tab:rdkit_classification_descriptors_b} report the top 15 RDKit descriptors for classification tasks. Table~\ref{tab:rdkit_regression_descriptors} reports the top 15 RDKit descriptors for regression tasks.

\begin{table}[H]
\centering
\scriptsize
\setlength{\tabcolsep}{2.5pt}
\resizebox{\columnwidth}{!}{
\begin{tabular}{clll}
\toprule
Rank & BACE & BBBP & ClinTox \\
\midrule
1 & NumHeteroatoms & TPSA & BalabanJ \\
2 & BertzCT & NOCount & NumAromaticHeterocycles \\
3 & HeavyAtomCount & NumHDonors & fr\_quatN \\
4 & Chi0 & NumHeteroatoms & fr\_NH0 \\
5 & Chi1 & NHOHCount & fr\_Ar\_N \\
6 & SlogP\_VSA2 & NumHAcceptors & fr\_aniline \\
7 & MaxEStateIndex & VSA\_EState3 & NumAromaticRings \\
8 & MaxAbsEStateIndex & PEOE\_VSA10 & MaxPartialCharge \\
9 & HeavyAtomMolWt & EState\_VSA10 & SMR\_VSA3 \\
10 & NumValenceElectrons & PEOE\_VSA1 & BCUT2D\_LOGPLOW \\
11 & ExactMolWt & qed & BCUT2D\_MRLOW \\
12 & MolWt & EState\_VSA1 & PEOE\_VSA8 \\
13 & LabuteASA & SMR\_VSA1 & SlogP\_VSA8 \\
14 & SMR\_VSA1 & fr\_lactam & SlogP\_VSA10 \\
15 & Chi0n & SlogP\_VSA2 & PEOE\_VSA3 \\
\bottomrule
\end{tabular}
}
\caption{Top 15 RDKit descriptors for BACE, BBBP, and ClinTox.}
\label{tab:rdkit_classification_descriptors_a}
\end{table}

\begin{table}[H]
\centering
\scriptsize
\setlength{\tabcolsep}{2.5pt}
\resizebox{\columnwidth}{!}{
\begin{tabular}{clll}
\toprule
Rank & HIV & Tox21 & SIDER \\
\midrule
1 & fr\_azo & NumAromaticCarbocycles & AvgIpc \\
2 & NumHeteroatoms & fr\_benzene & FpDensityMorgan3 \\
3 & TPSA & NumAromaticRings & MaxAbsEStateIndex \\
4 & BertzCT & SMR\_VSA7 & MaxEStateIndex \\
5 & HeavyAtomMolWt & SlogP\_VSA6 & BCUT2D\_CHGHI \\
6 & ExactMolWt & SlogP\_VSA8 & BCUT2D\_LOGPHI \\
7 & MolWt & Chi3v & NumAromaticRings \\
8 & NOCount & RingCount & fr\_ketone \\
9 & SMR\_VSA10 & BertzCT & NumAliphaticCarbocycles \\
10 & PEOE\_VSA13 & Chi4v & fr\_NH0 \\
11 & LabuteASA & fr\_phenol\_noOrthoHbond & RingCount \\
12 & Kappa1 & VSA\_EState6 & NumSaturatedCarbocycles \\
13 & Chi0 & fr\_phenol & qed \\
14 & HeavyAtomCount & NumAliphaticCarbocycles & fr\_piperzine \\
15 & NumValenceElectrons & NumSaturatedCarbocycles & PEOE\_VSA3 \\
\bottomrule
\end{tabular}
}
\caption{Top 15 RDKit descriptors for HIV, Tox21, and SIDER.}
\label{tab:rdkit_classification_descriptors_b}
\end{table}

\begin{table}[H]
\centering
\scriptsize
\setlength{\tabcolsep}{2.5pt}
\resizebox{\columnwidth}{!}{
\begin{tabular}{clll}
\toprule
Rank & ESOL & FreeSolv & Lipo \\
\midrule
1 & MolLogP & TPSA & MolLogP \\
2 & PEOE\_VSA6 & NumHDonors & fr\_COO \\
3 & MolMR & NOCount & fr\_COO2 \\
4 & Chi0v & NHOHCount & NumAromaticRings \\
5 & LabuteASA & MinPartialCharge & RingCount \\
6 & MolWt & PEOE\_VSA1 & fr\_Al\_COO \\
7 & HeavyAtomMolWt & NumHAcceptors & BalabanJ \\
8 & ExactMolWt & MaxAbsPartialCharge & FpDensityMorgan1 \\
9 & Chi1v & SlogP\_VSA2 & AvgIpc \\
10 & BCUT2D\_LOGPHI & MaxAbsEStateIndex & SlogP\_VSA6 \\
11 & Chi2v & MaxEStateIndex & VSA\_EState6 \\
12 & FpDensityMorgan1 & VSA\_EState2 & SMR\_VSA7 \\
13 & Chi1 & VSA\_EState3 & NHOHCount \\
14 & HeavyAtomCount & SMR\_VSA3 & BertzCT \\
15 & Chi3v & MinAbsPartialCharge & EState\_VSA7 \\
\bottomrule
\end{tabular}
}
\caption{Top 15 RDKit descriptors for regression tasks. Descriptors were ranked by absolute Pearson correlation with the target value.}
\label{tab:rdkit_regression_descriptors}
\end{table}

\section{Results for Multitask Datasets}
\label{app:per-task}

Tables~\ref{tab:multitask_clintox}, \ref{tab:multitask_tox21}, and~\ref{tab:multitask_sider} report results for the multitask binary classification datasets. We report ROC-AUC for the SMILES-only baseline and the full \molerag setting. These results summarize performance over the available task labels within each dataset.

Across multitask datasets, \molerag generally improves ROC-AUC over the SMILES-only baseline for most models. The gains are especially clear on ClinTox, where nearly all evaluated models improve with \molerag; for example, Mistral increases from $0.500$ to $0.621$, ChemDFM from $0.497$ to $0.594$, GPT-4o-mini from $0.520$ to $0.633$, and GPT-5.4-nano from $0.528$ to $0.627$. On Tox21, \molerag also improves most evaluated models, with particularly large gains for Qwen, ChemDFM, and GPT-5.4-nano. SIDER shows more modest gains overall, although GPT-5.4-nano, ChemDFM, GPT-4o-mini, and Qwen still improve relative to the SMILES-only baseline. These results suggest that multitask datasets benefit from inference-time molecular context, but the magnitude of improvement depends on both the model and the specific set of assay labels.

\begin{table}[H]
\centering
\scriptsize
\setlength{\tabcolsep}{3pt}
\resizebox{\columnwidth}{!}{
\begin{tabular}{lcc}
\toprule
Model & SMILES ROC-AUC & \molerag ROC-AUC \\
\midrule
Llama-3.2-3B-Instruct & $0.483$ & $0.488$ \\
Mistral-7B-Instruct-v0.3 & $0.500$ & $0.621$ \\
Qwen3-4B-Instruct-2507 & $0.503$ & $0.566$ \\
ChemDFM-v2.0-14B & $0.497$ & $0.594$ \\
GPT-4o-mini & $0.520$ & $0.633$ \\
GPT-5.4-nano & $0.528$ & $0.627$ \\
\bottomrule
\end{tabular}
}
\caption{ROC-AUC results on ClinTox.}
\label{tab:multitask_clintox}
\end{table}

\begin{table}[H]
\centering
\scriptsize
\setlength{\tabcolsep}{3pt}
\resizebox{\columnwidth}{!}{
\begin{tabular}{lcc}
\toprule
Model & SMILES ROC-AUC & \molerag ROC-AUC \\
\midrule
Llama-3.2-3B-Instruct & $0.500$ & $0.525$ \\
Mistral-7B-Instruct-v0.3 & $0.531$ & $0.548$ \\
Qwen3-4B-Instruct-2507 & $0.496$ & $0.662$ \\
ChemDFM-v2.0-14B & $0.550$ & $0.688$ \\
GPT-4o-mini & $0.523$ & $0.593$ \\
GPT-5.4-nano & $0.527$ & $0.670$ \\
\bottomrule
\end{tabular}
}
\caption{ROC-AUC results on Tox21.}
\label{tab:multitask_tox21}
\end{table}

\begin{table}[H]
\centering
\scriptsize
\setlength{\tabcolsep}{3pt}
\resizebox{\columnwidth}{!}{
\begin{tabular}{lcc}
\toprule
Model & SMILES ROC-AUC & \molerag ROC-AUC \\
\midrule
Llama-3.2-3B-Instruct & -- & $0.503$ \\
Mistral-7B-Instruct-v0.3 & $0.504$ & $0.508$ \\
Qwen3-4B-Instruct-2507 & $0.511$ & $0.528$ \\
ChemDFM-v2.0-14B & $0.492$ & $0.539$ \\
GPT-4o-mini & $0.516$ & $0.535$ \\
GPT-5.4-nano & $0.516$ & $0.566$ \\
\bottomrule
\end{tabular}
}
\caption{ROC-AUC results on SIDER.}
\label{tab:multitask_sider}
\end{table}

\section{Usage of AI Assistants}
\label{app:ai-usage}

In preparing this work, we used AI-based writing assistants to improve sentence structure, correct grammatical errors, and enhance overall readability. These tools were used only for language editing and presentation refinement. They did not determine the research questions, experimental design, methodology, implementation, analysis, or conclusions of the paper. All scientific claims, experimental results, and interpretations were reviewed and verified by the authors.

We also used large language models (LLMs) as part of the experimental pipeline. Specifically, GPT-4o-mini was used to generate task-specific keywords for the hybrid textual retrieval queries. These generated keywords were cached and reused across experiments to improve lexical overlap with chemistry-related passages. In addition, several LLMs, including proprietary, open-source, and chemistry-specialized models, were evaluated as prediction models in our benchmark. Their use as experimental subjects is described in the main method and experiment sections.

No AI assistant was used to fabricate data, alter experimental results, or make autonomous scientific decisions. All generated text, prompts, retrieval outputs, and model predictions were inspected, processed, and analyzed by the authors according to the experimental protocol.

\section{Scientific Artifacts}
\label{app:artifacts}

This work uses publicly available datasets, software libraries, pretrained models, and retrieval corpora. All artifacts were used in a manner consistent with their intended research purposes.

\paragraph{Datasets}
We evaluate on molecular property prediction tasks from MoleculeNet \citep{moleculenet}, including BBBP, BACE, ClinTox, HIV, Tox21, SIDER, ESOL, FreeSolv, and Lipophilicity. These datasets are standard benchmarks for molecular property prediction and were used only for evaluating model performance under scaffold-based train/validation/test splits.

\paragraph{Retrieval corpus}
For textual retrieval, we use the ChemRAG corpus \citep{Zhong-ChemRAG}, which aggregates chemistry-related text from resources such as PubChem, PubMed, USPTO, Semantic Scholar, Wikipedia, and OpenStax textbooks. We use this corpus only as an external retrieval source for inference-time textual evidence.

\paragraph{Software tools}
We use RDKit to compute molecular descriptors, molecular fingerprints, Tanimoto similarity, and other cheminformatics features. RDKit \citep{landrum2013rdkit} was used for molecular structure analysis and descriptor computation, consistent with its intended use. We also use AccFG \citep{liu2025accfg} for functional group annotation, which provides molecule-level functional group information included in the molecular context prompt.

\paragraph{Models}
We evaluate proprietary, open-source, and chemistry-specialized LLMs, including GPT-4o-mini \citep{openai2024gpt4omini}, GPT-5.4-nano \citep{singh2025openai}, Llama-3.2-3B-Instruct \citep{llama}, Mistral-7B-Instruct-v0.3 \citep{jiang2023mistral}, Qwen3-4B-Instruct \citep{yang2025qwen3}, ChemLLM-7B-Chat \citep{zhang2024chemllm}, and ChemDFM-v2.0-14B \citep{ChemDFM2025}. These models were used as inference-time predictors and were not finetuned in this work. All models were evaluated under the same prompt-based framework with temperature set to $0$.

\paragraph{Package details}
All experiments were implemented in Python~3.10.12. LLM inference for local models used PyTorch~2.8.0 with CUDA~12.8, Hugging Face Transformers~4.57.1, Accelerate~1.13.0, Datasets~2.19.2, and Tokenizers~0.22.1. Molecular preprocessing, descriptor computation, and fingerprint-based similarity search used RDKit~2026.03.2. Functional group annotations were generated using AccFG. BM25 retrieval was implemented using \texttt{rank\_bm25}. Data processing used NumPy~2.2.6 and pandas~2.3.3. API-based model calls used the OpenAI Python package~2.38.0. We also used SentencePiece~0.2.1 and Protobuf~7.35.0 for tokenizer and model-loading support.

\paragraph{License and intended use}
All datasets, corpora, software tools, and pretrained models were used according to their respective licenses and terms of use. The artifacts were used for molecular property prediction, retrieval, descriptor computation, and benchmarking, which are consistent with their intended research applications. No dataset was used to identify individuals or make clinical decisions.

\section{Prompt Templates}
\label{app:prompts}

This section provides representative prompt templates used in our experiments. To fit the two-column format, we present compact templates that preserve the main instruction structure and output constraints. Context blocks were included or omitted depending on the experimental condition. The SMILES-only baseline used only the task instruction and query molecule. The single-component ablation prompts isolate textual retrieval, molecular context, and structure-based retrieval, respectively. The full \molerag prompt combines retrieved chemistry passages, molecule-specific context, and structurally similar labeled molecules. We also include the synonym-filtering prompt used to construct the filtered synonym cache.

\begin{tcolorbox}[
  breakable,
  width=\columnwidth,
  title={Prompt 1: SMILES-Only Classification},
  colback=gray!10,
  colframe=black!60,
  fonttitle=\bfseries,
  left=1mm,
  right=1mm,
  top=1mm,
  bottom=1mm
]
\refstepcounter{promptbox}\label{prompt:smiles-classification}
\begin{Verbatim}[breaklines, fontsize=\scriptsize]
System:
You are a chemistry expert.

User:
Task: Predict [TASK_DESCRIPTION].

SMILES: [QUERY_SMILES]

Reply with EXACTLY ONE WORD:
either 'Yes' or 'No'.
Output nothing else. No explanation,
no punctuation, no context.
\end{Verbatim}
\end{tcolorbox}

\begin{tcolorbox}[
  breakable,
  width=\columnwidth,
  title={Prompt 2: SMILES-Only Regression},
  colback=gray!10,
  colframe=black!60,
  fonttitle=\bfseries,
  left=1mm,
  right=1mm,
  top=1mm,
  bottom=1mm
]
\refstepcounter{promptbox}\label{prompt:smiles-regression}
\begin{Verbatim}[breaklines, fontsize=\scriptsize]
System:
You are a chemistry expert.

User:
Task: Predict [TASK_DESCRIPTION].

SMILES: [QUERY_SMILES]

Predict the [REGRESSION_PROPERTY].
Valid predictions are real numbers
[VALID_RANGE].

Output EXACTLY ONE number
(example: [EXAMPLE_VALUE]).
No units, no explanation.

ANSWER:
\end{Verbatim}
\end{tcolorbox}

\begin{tcolorbox}[
  breakable,
  width=\columnwidth,
  title={Prompt 3: Textual Retrieval Only},
  colback=gray!10,
  colframe=black!60,
  fonttitle=\bfseries,
  left=1mm,
  right=1mm,
  top=1mm,
  bottom=1mm
]
\refstepcounter{promptbox}\label{prompt:textual-only}
\begin{Verbatim}[breaklines, fontsize=\scriptsize]
System:
You are a chemistry expert.

User:
Task: Predict [TASK_DESCRIPTION].

Retrieved chemistry context:
[TOP-5_RETRIEVED_PASSAGES]

Molecule:
SMILES: [QUERY_SMILES]

Reply with EXACTLY ONE WORD:
either 'Yes' or 'No'.
Output nothing else. No explanation,
no punctuation, no context.
\end{Verbatim}
\end{tcolorbox}

\begin{tcolorbox}[
  breakable,
  width=\columnwidth,
  title={Prompt 4: Molecular Context Only},
  colback=gray!10,
  colframe=black!60,
  fonttitle=\bfseries,
  left=1mm,
  right=1mm,
  top=1mm,
  bottom=1mm
]
\refstepcounter{promptbox}\label{prompt:molecular-context-only}
\begin{Verbatim}[breaklines, fontsize=\scriptsize]
System:
You are a chemistry expert.

User:
Task: Predict [TASK_DESCRIPTION].

Compound identifiers:
Name: [COMMON_NAME]
Other names: [SYNONYMS]
IUPAC: [IUPAC_NAME]

Functional groups:
[FUNCTIONAL_GROUPS]

RDKit descriptors:
[DESCRIPTOR_1]: [VALUE_1]
...
[DESCRIPTOR_15]: [VALUE_15]

Lipinski Rule of 5:
MolWt=[VALUE], MolLogP=[VALUE],
HBD=[VALUE], HBA=[VALUE],
violations=[VALUE]

Molecule:
SMILES: [QUERY_SMILES]

Reply with EXACTLY ONE WORD:
either 'Yes' or 'No'.
Output nothing else. No explanation,
no punctuation, no context.
\end{Verbatim}
\end{tcolorbox}

\begin{tcolorbox}[
  breakable,
  width=\columnwidth,
  title={Prompt 5: Structure Retrieval Only},
  colback=gray!10,
  colframe=black!60,
  fonttitle=\bfseries,
  left=1mm,
  right=1mm,
  top=1mm,
  bottom=1mm
]
\refstepcounter{promptbox}\label{prompt:structure-only}
\begin{Verbatim}[breaklines, fontsize=\scriptsize]
System:
You are a chemistry expert.

User:
Task: Predict [TASK_DESCRIPTION].

Structurally similar training molecules:
Example 1: SMILES=[TRAIN_SMILES_1],
label=[LABEL_1], sim=[SIM_1]
...
Example 5: SMILES=[TRAIN_SMILES_5],
label=[LABEL_5], sim=[SIM_5]

Query molecule:
SMILES: [QUERY_SMILES]

Reply with EXACTLY ONE WORD:
either 'Yes' or 'No'.
Output nothing else. No explanation,
no punctuation, no context.
\end{Verbatim}
\end{tcolorbox}

\begin{tcolorbox}[
  breakable,
  width=\columnwidth,
  title={Prompt 6: Full \molerag Classification},
  colback=gray!10,
  colframe=black!60,
  fonttitle=\bfseries,
  left=1mm,
  right=1mm,
  top=1mm,
  bottom=1mm
]
\refstepcounter{promptbox}\label{prompt:full-molerag-classification}
\begin{Verbatim}[breaklines, fontsize=\scriptsize]
System:
You are a chemistry expert.

User:
Task: Predict [TASK_DESCRIPTION].

Retrieved chemistry context:
[TOP-5_RETRIEVED_PASSAGES]

Structurally similar training molecules:
Example 1: SMILES=[TRAIN_SMILES_1],
label=[LABEL_1], sim=[SIM_1]
...
Example 5: SMILES=[TRAIN_SMILES_5],
label=[LABEL_5], sim=[SIM_5]

Compound identifiers:
Name: [COMMON_NAME]
Other names: [SYNONYMS]
IUPAC: [IUPAC_NAME]

Functional groups:
[FUNCTIONAL_GROUPS]

RDKit descriptors:
[DESCRIPTOR_1]: [VALUE_1]
...
[DESCRIPTOR_15]: [VALUE_15]

Lipinski Rule of 5:
MolWt=[VALUE], MolLogP=[VALUE],
HBD=[VALUE], HBA=[VALUE],
violations=[VALUE]

Now predict the answer for this molecule.
SMILES: [QUERY_SMILES]

Reply with EXACTLY ONE WORD:
either 'Yes' or 'No'.
Output nothing else.
\end{Verbatim}
\end{tcolorbox}

\begin{tcolorbox}[
  breakable,
  width=\columnwidth,
  title={Prompt 7: Full \molerag Regression},
  colback=gray!10,
  colframe=black!60,
  fonttitle=\bfseries,
  left=1mm,
  right=1mm,
  top=1mm,
  bottom=1mm
]
\refstepcounter{promptbox}\label{prompt:full-molerag-regression}
\begin{Verbatim}[breaklines, fontsize=\scriptsize]
System:
You are a chemistry expert.

User:
Task: Predict [TASK_DESCRIPTION].

Retrieved chemistry context:
[TOP-5_RETRIEVED_PASSAGES]

Structurally similar training molecules:
Example 1: SMILES=[TRAIN_SMILES_1],
true value=[VALUE_1], sim=[SIM_1]
...
Example 5: SMILES=[TRAIN_SMILES_5],
true value=[VALUE_5], sim=[SIM_5]

Compound identifiers:
Name: [COMMON_NAME]
Other names: [SYNONYMS]
IUPAC: [IUPAC_NAME]

Functional groups:
[FUNCTIONAL_GROUPS]

RDKit descriptors:
[DESCRIPTOR_1]: [VALUE_1]
...
[DESCRIPTOR_15]: [VALUE_15]

Target SMILES: [QUERY_SMILES]

Predict the [REGRESSION_PROPERTY]
for THIS target molecule.
Valid predictions are real numbers
[VALID_RANGE].

CRITICAL: Descriptor values and
molecular identifiers are properties of
the molecule, NOT the answer. Do not
echo those numbers.

Output EXACTLY ONE number
(example: [EXAMPLE_VALUE]).
No units, no descriptor names,
no explanation.

ANSWER:
\end{Verbatim}
\end{tcolorbox}

\begin{tcolorbox}[
  breakable,
  width=\columnwidth,
  title={Prompt 8: Synonym Cache Filtering},
  colback=gray!10,
  colframe=black!60,
  fonttitle=\bfseries,
  left=1mm,
  right=1mm,
  top=1mm,
  bottom=1mm
]
\refstepcounter{promptbox}\label{prompt:synonym-cache-filtering}
\begin{Verbatim}[breaklines, fontsize=\scriptsize]
System:
You output JSON only. No prose,
no markdown.

User:
You are filtering chemical compound
synonyms for a literature search.

Keep paper-friendly names, such as
drug names, trade names, trivial names,
and IUPAC-style descriptive names.

Drop catalog, registry, database,
supplier, CAS, and lab notebook codes,
such as NSC IDs, CHEMBL IDs, NCI60 IDs,
SCHEMBL IDs, AKOS IDs, and CAS numbers.

Return only the paper-friendly names.
Order them by how commonly they appear
in chemistry literature.

IUPAC name: [IUPAC_NAME]
Candidate synonyms: [CANDIDATE_SYNONYMS]

Respond with JSON only:
{"keep": []}

Use an empty list if none are
paper-friendly.
\end{Verbatim}
\end{tcolorbox}

\end{document}